\def\year{2020}\relax
\documentclass[letterpaper]{article} 
\usepackage{aaai20}  
\usepackage{times}  
\usepackage{helvet} 
\usepackage{courier}  
\usepackage[hyphens]{url}  
\usepackage{graphicx} 
\usepackage{graphicx}  
\usepackage{array}

\usepackage[finalnew]{trackchanges}

\addeditor{KV}
\addeditor{JL}

\usepackage{subcaption}

\copyrighttext{%
  Distribution A: Approved for public release, distribution unlimited.\\
}

\title{%
  Data-Driven Approaches for Thrust Prediction in \\ Underwater Flapping Fin Propulsion Systems
}
\author{%
     Julian Lee,\textsuperscript{\rm 1}\thanks{julian.lee@yale.edu}
     Kamal Viswanath,\textsuperscript{\rm 2}\thanks{kamal.viswanath@nrl.navy.mil}
     Alisha Sharma,\textsuperscript{\rm 2} \\
     \textbf{\Large{Jason Geder,\textsuperscript{\rm 2}
    Ravi Ramamurti,\textsuperscript{\rm 2}
    Marius D. Pruessner.\textsuperscript{\rm 2}}}\\
    \textsuperscript{\rm 1}Yale University, New Haven, CT 06520\\
    \textsuperscript{\rm 2}Naval Research Laboratory, 4555 Overlook Ave SW, Washington, DC, 20375\\
}

\begin{document}

\maketitle

\begin{abstract}
  Flapping-fin underwater vehicle propulsion systems provide an alternative to propeller-driven systems in situations that involve a
  constrained environment or require high maneuverability. Testing new
  configurations through experiments or high-fidelity simulations is
  an expensive process, slowing development of new systems. This is especially
  true when introducing new fin geometries.
  In this work, we propose machine learning approaches for thrust prediction given the system’s fin geometries and kinematics. We introduce data-efficient fin shape parameterization strategies that enable our network to predict thrust profiles for unseen fin geometries given limited fin shapes in input data. In addition to faster development of systems, generalizable surrogate models offer fast, accurate predictions that could be used on an unmanned underwater vehicle control system. 

\end{abstract}

\section{Introduction}

\noindent Bio-inspired flapping-fin propulsion systems are an alternative
to traditional propeller-driven systems for unmanned underwater vehicle (UUV) operations.
Unlike propeller-driven systems, which are often optimized for propulsive efficiency,
bio-inspired systems perform well in constrained or unsteady environments (e.g.
near-shore regions) and are high maneuverable at low speeds \cite{Lic04}. Various studies have targeted the use of flapping fins to achieve thrust, with a focus on studying fin shapes and materials as well as stroke kinematic parameters and fin configurations \cite{Hobson,Barrett,Kato}. 

However, designing these systems are costly: high-fidelity computational
fluid dynamics (CFD) used in previous work \cite{Muscutt,Ramamurti_Computational,Ramamurti_Fluid} takes on the order of hours for a single simulation while experimental rigs takes days to set up and run.
This poses a challenge for exploring the design space of flapping fin systems and
developing model-driven adaptive vehicle control systems. 

Two common alternatives to full CFD are single-point models, which model a single configuration for some desired fidelity, and quasi-steady reduced-order models \cite{Geder_Four}, which use approximate representations of the underlying physics for fast, low-fidelity predictions. However, both approaches require steep trade-offs in model generalizability and/or fidelity.

Due to increasing amounts of available data and computational power as well as recent theoretical developments, data-driven surrogate modeling has become increasingly popular. Through the use of machine learning, we can produce a model that does not contain explicitly embedded mathematics. Additionally, compared to CFD approaches \cite{Ramamurti_Computational}, neural networks offer the benefit of an efficient runtime with a relatively large parameter space, minimizing the need for a highly reduced order model and thus potentially improving accuracy.

Flapping-fin propulsion systems have many salient design features,
but one of the most challenging to represent is the fin geometry (including
taper, planform shape, and cross-sectional shape).
This design choice impacts the propulsive performance of the system
and thus is important if we want to accurately model the system behavior.  

\begin{figure}[t]
  \centering
	\includegraphics[width=0.8\linewidth]{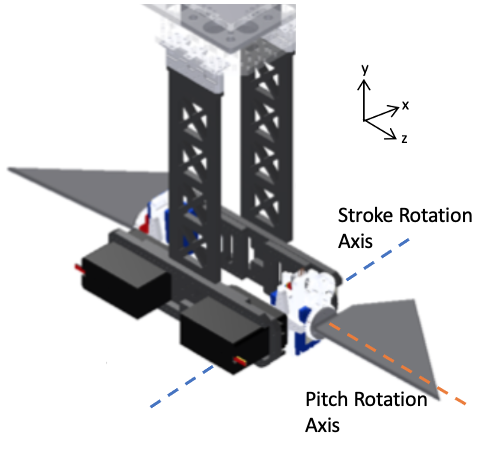}
	\caption{
    Example simple flapping fin propulsion system composed of paired pectoral (side) fins.
	}
	\label{fig:config}
\end{figure}

In this study, we design surrogate models incorporating fin parameterization to predict
propulsive performance given fin geometry and system kinematics. Due to the expense of testing new fin geometries, we aim to develop data-efficient surrogate models \cite{Viswanath} that allow for generalizability to new geometries despite limited fin shapes provided in building our model.

\section{Methodology}

\subsection{Neural Network Parameters}

Our proposed models use both geometry-related parameters as well as stroke-related parameters to predict the thrust of a paired pectoral fin propulsion system. Geometry-related parameters refer to the construction of the UUV fins while stroke-related parameters refer to the UUV’s current kinematics and often vary as a function of time. \add[KV]{For the flapping fin setup illustrated in} Fig. \ref{fig:config}, \add[KV]{the incoming flow is in the postive x-direction and the straight edge of the fin, parallel to the pitch rotation axis, on which the flow is first incident is termed the "leading edge" of the fin and the flows leaves from the "trailing edge". A "chord" on the fin is any cross-section cut from the leading edge to the trailing edge in the direction of the flow. The "root" of the fin is where it attaches to the UUV body, or the motor in} Fig. \ref{fig:config}{,  and conversely the farthest point of the fin from the root is the "fin tip". For the kinematics, a "stroke cycle" describes an upstroke and a downstroke going through the full excursion of the fin tip and returning to the same kinematics state at which it started about the stroke rotation axis. }A list describing the parameters for thrust prediction is shown in Table \ref{tab:parameters}.

\begin{table}[ht]
\footnotesize
\centering
 \caption{Input parameters for neural networks predicting thrust generated by pectoral fins}
 \label{tab:parameters}
 \begin{tabular}{lp{4.5cm}}
 \hline
 \hline
 Parameter & Description \\  
 \hline
 \hline
 \textit{Geometry-related} &  \\ 
 \hline
 Fin Geometry & Quantification of fin shape (described in “Representing Fin Geometries”) \\
 \hline
 \textit{Stroke-related} & \\
 \hline
 Stroke Amplitude (°) & Maximum stroke angle reached during a stroke cycle \\
 \hline
 Pitch Amplitude (°) & Maximum pitch angle reached during a stroke cycle \\
 \hline
 Flap Frequency (Hz) & Frequency of a single stroke cycle \\
 \hline
 Stroke Angle (°) & History of the flapping angle over the course of a stroke cycle as a function of time \\
 \hline
 Pitch Angle (°)	& History of the pitching angle over the course of a stroke cycle as a function of time \\
 \hline
 Stroke state & Binary value set to 1 during upstroke and 0 during downstroke
\end{tabular}
\end{table}%

\subsection{Representing Fin Geometries}

When designing a surrogate model for flapping fin propulsion, we must determine how to represent our geometry to the network. Passing fin shape as a single categorical value to the neural network results in a model will be ineffective in generalizing to new fin shapes. Providing quantitative information about the given fin shape allows the neural network to build an understanding of how the fin geometry influences the generated thrust. Therefore, we quantify fin shapes as a series of parameters that can be used by the neural network to inform thrust prediction. 

In a previous study, Samareh \shortcite{Sam99} identified eight broad approaches
to shape parameterization for airfoils: basis vectors, domain element approaches,
partial differential equations (PDEs), discrete representations,
polynomials/splines, computer-aided design (CAD) representations, analytical
representations, and free-form deformation (FFD) \cite{Sam99}. We investigate how discrete representations can lead to an understanding of how fin shape influences thrust for our neural networks. 

\begin{figure}[t]
  \centering
	\includegraphics[width=0.6\linewidth]{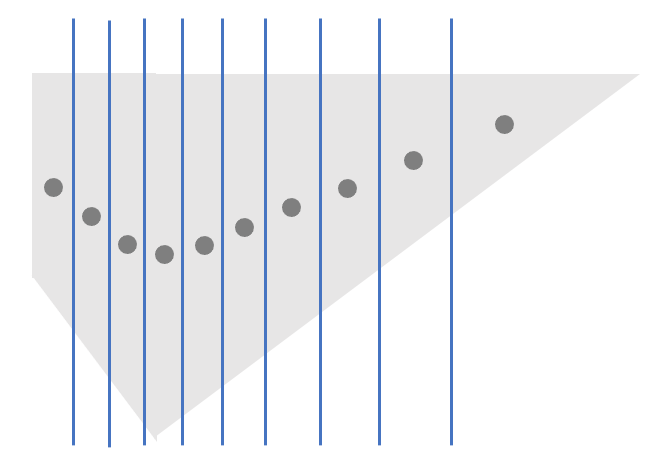}
	\caption{
    Example division of flat fin into 10 equal-area regions. Resulting center of masses shown in gray.
	}
	\label{fig:segmentation}
\end{figure}

For the discrete spatial representation, the frame of reference origin is fixed at the intersection of the stroke axis and the pitch axis. To parameterize fin shapes, each fin is divided into 10 equal-area regions using lines that are parallel to the stroke axis. Segments are chosen in this manner since a given fin segment's area contribution to thrust is dependent on its distance from the root of the fin. The center of mass for each segment is then computed for the flat fins. These points each represent an equal-area component of the fin, and they collectively form a simplified “skeletal” model of the given geometry (Fig \ref{fig:segmentation}). 

To facilitate a learned relation between the fin’s geometry and its current position in 3D space, the 2D centers of mass are rotated about the stroke and pitch rotation axes by the current stroke and pitch angles. Through this transformation, the resulting 30 values consisting of ten 3D centers of mass capture the instantaneous shape and position of the fin. 

If only the flat fin was considered for fin parameterization, these values would remain static for each given fin shape, leading to sparse domain coverage from the input data. Considering the 3D location of the fin in the calculation of fin shape parameters allows for dynamic values that vary as the flapping cycle progresses or if the kinematic settings are changed. Therefore, such a variance in inputs can lead to a more data-efficient network that requires fewer fin shape geometries during training.

\subsubsection{Reduced-Order Fin Parameterization}

We investigate reducing the fin geometry input space from 30 parameters to fewer parameters through the use of principal component analysis. Fin shape parameters are calculated based on the kinematic-shape settings present in our experimental data (Section 3), and PCA is applied on this 30-dimensional space. Prior to PCA, a weighted and non-weighted scaling method is proposed. In the non-weighted procedure, each dimension was scaled to have an equal standard deviation. In the weighted procedure, each dimension was scaled such that the standard deviation of values across all data was equal to the Pearson correlation between the given parameter and the normalized thrust coefficient. 

Such a weighting procedure puts a greater emphasis on the inputs that have a clear correlation with thrust, allowing these values to have a stronger influence on the values generated by the PCA. Across our experimental data explained in Section 3, parameters representing the COM locations in the X and Z dimensions had a larger linear association with thrust compared with the parameters representing the COM locations in the physical Y dimension (Table 6), so inputs representing values in the X and Z dimensions are potentially more valuable in thrust prediction. Since the flat fin lies on the xz plane, this emphasis on the x and z plane could also result in a clearer differentiation between fin shapes through the reduced-order fin parameterization. 

With either method, almost all of the variance in our fin shape parameters can be explained using relatively few principal components. The ratio between the standard deviation in the first and fourth principal component is 9.39 for the weighted method and 97.2 for the unweighted method----the vast majority of the parameter space can be captured with four parameters.

\subsection{Neural Network Architectures}

We experiment with two architectures for thrust prediction: dense neural networks and long-term short memory neural networks. The dense neural network architecture contains layers of nodes such that each node is connected to every node in both the previous and subsequent layer. Our dense neural network aims to make instantaneous thrust predictions: given the current stroke and pitch angle as well as the other parameters specified in Table \ref{tab:parameters}, the network will output a single value denoting the predicted thrust at that instant in time. 

Since the proposed dense neural network predicts thrust at a specific point in time, the network cannot build a memory of the fin kinematics at prior points in time to improve its predictions. Therefore, we propose the use of a long-term short memory neural network \add[KV]{(LSTM)}, a type of recurrent neural network. Recurrent neural networks process time-series data, and they use the predicted outputs from the previous time step to help predict the output in the current time step. LSTMs additionally maintain a cell state that stores information from multiple past time steps \add[KV]{to inform the current one.}  

We will provide the kinematics of an entire stroke cycle as a training example for our LSTM, and our LSTM will then output a thrust profile. We propose a many-to-many LSTM approach (Figure 6), which involves generating a hidden state from each input and using these hidden states to calculate thrust.

\section{Experimental Setup}

\subsection{The Dataset}

We used experimental data collected by the Laboratories for Computational Physics \& Fluid Dynamics. During experimental trials, the stroke rotation axis has an offset 3.175 cm from the root chord of the fin, and the pitch rotation axis has an offset of 1.25 cm from the fin leading edge. The data consisted of two flap frequencies (1 Hz and 2 Hz), five pitch amplitudes (0°, 15°, 25°, 40°, and 55°), three fin geometries (rectangular, bio, pt4). The stroke amplitude was set to 60° for the test runs with a flap frequency of 1 Hz, and the stroke amplitude was set to 25° for the 2 Hz runs. Five flapping cycles were provided for each experiment, and 16 experimental runs were provided for each kinematic-shape setting. Data from the 2 Hz, 25° pitch amplitude bio runs were excluded due to a sensor malfunction during this run.

\begin{figure}[t]
  \centering
	\includegraphics[width=0.8\linewidth]{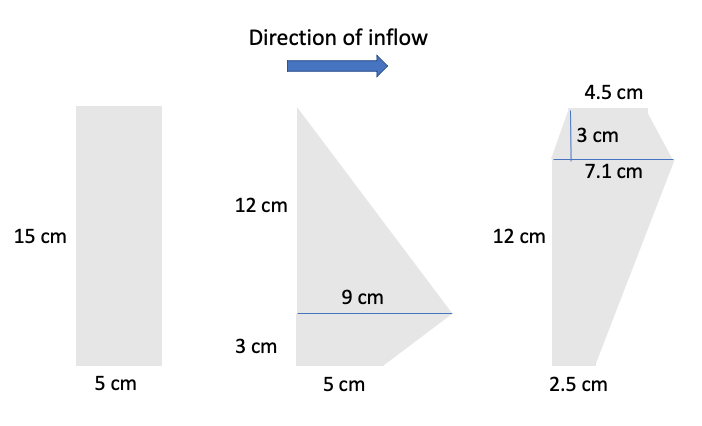}
	\caption{
    Experimental data fin geometries for the rectangular fin (left), bio fin (middle), and pt4 fin (right) 
	}
	\label{fig:fin_geometries}
\end{figure}

\begin{figure}[t]
  \centering
	\includegraphics[width=0.6\linewidth]{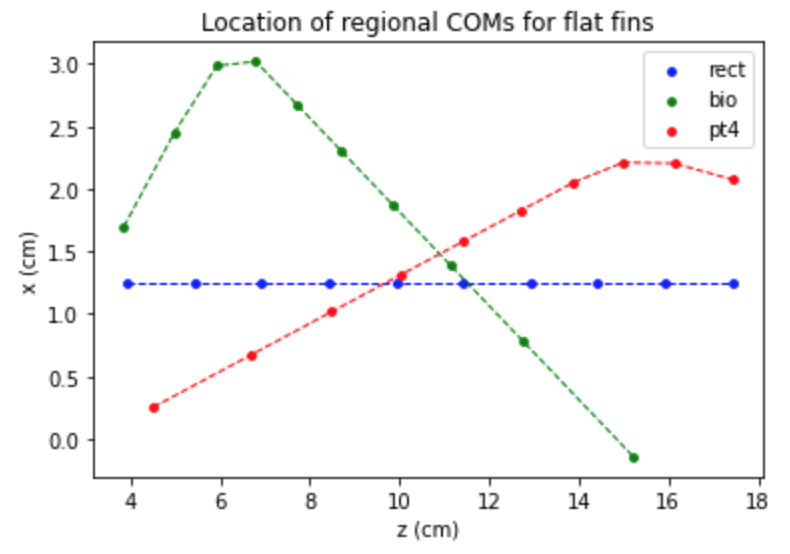}
	\caption{
    Location of regional center of masses for flat fins
	}
	\label{fig:fin_COMs}
\end{figure}

The three fin geometries included in the data are shown in Figure \ref{fig:fin_geometries}, and the location of the flat fin center of masses for these geometries are shown in Figure \ref{fig:fin_COMs}. The rectangular fin serves as a baseline synthetic geometry while the bio fin and pt4 fin draw inspiration from nature. The bio fin was created through the study of fish pectoral fins while insect wings were used to motivate the pt4 fin design.

\subsection{Data Pre-Processing}

Z-score normalization was applied for each input parameter. To normalize thrust, we computed the thrust coefficient as  $T_{coeff} = \frac{T}{\frac{1}{2}\rho v^2 A}$, where $\rho$ represents the material property, v represents our reference velocity (in this case the average fin tip velocity), and l represents the reference area (in this case the area of the fin). The thrust coefficients are then linearly scaled such that their standard deviation equals 1. 


\subsection{Data Noise Evaluation}

\begin{figure}[t]
  \centering
	\includegraphics[width=0.6\linewidth]{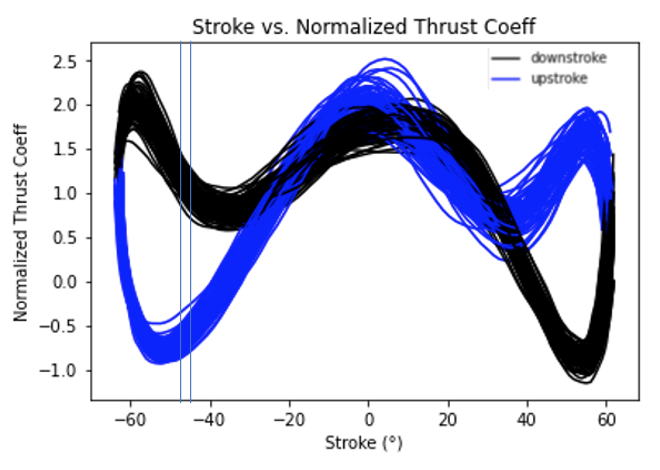}
	\caption{
    Sample experimental data for a kinematic-shape setting. A sample interval is shown in between the blue bars.
	}
	\label{fig:noise}
\end{figure}

The provided data set contains 80 flapping cycles across 16 experimental runs for each kinematic-shape setting. We created a thrust deviation metric to quantify the noise within each group of 80 flapping cycles. The metric is calculated as follows: Define the stroke range for a given flapping cycle group as [$s_{min}$,$s_{max}$], where $s_{min}$ is the minimum stroke angle in the stroke cycle and $s_{max}$ is the maximum stroke angle. The stroke range is divided into 100 equal-sized intervals, and data is also separated by upstroke and downstroke (Figure \ref{fig:noise}). Denoting the set of normalized thrust coefficients associated with interval i during upstroke as $ThrustCoeff_i$ and the set of normalized thrust coefficients associated with interval i during downstroke as $ThrustCoeff_{i+100}$, we can then compute the noise for each run as seen in $ThrustDev = \frac{\sum\limits_{i=1}^{200} \sigma(T_{CoeffNorm_i})}{200}$.

The mean thrust deviation across all our kinematic-shape settings was 0.2588, which yields a mean thrust variance of approximately 0.067. Therefore, a mean squared error of 0.067 generated by our neural networks is accounted by noise. Individually computing thrust deviation for each experimental run yielded a lower mean thrust deviation of 0.103. This lower deviation demonstrates a substantially higher variance in thrust profiles between different experimental runs compared with thrust profiles from the same experimental run. To create a representative set of data and incorporate the noise between experimental runs into our surrogate model training, one flapping cycle from each experimental run will be included in our reduced dataset that will be used for network training and validation.

\subsection{Generalizability Testing}

To test the generalizability of the model to interpolated kinematics, subsets of the kinematic data were excluded in the training of models as shown in Table \ref{tab:genDesc}. Gen test 1 excludes two unrelated kinematic-shape settings while gen tests 2-4 exclude all kinematic cases with the same pitch amplitude to create a harder interpolation problem. Gen tests 5 and 6 exclude all data from a either the rectangular or bio fin to test the model’s ability to generalize to new geometries. These tests exclude kinematic tests with a flap frequency of 2 Hz since this data is not included for the Pt4 fin. 

\begin{table}[ht]
\footnotesize
\centering
 \caption{Generalizability test descriptions}
 \label{tab:genDesc}
 \begin{tabular}{llp{3 cm}}
 \hline
 \hline
 Test Name & Evaluated Fin Shapes & Kinematics/Shapes \newline Excluded \\  
 \hline
 Gen Test 1 & Rect, Bio & Pitch amp = 40°, flap freq = 2 Hz, rect \newline Pitch amp = 15°, flap freq = 1 Hz, bio \\
 \hline
 Gen Test 2 & Rect, Bio & All pitch amp = 15° kinematic cases \\
 \hline
 Gen Test 3 & Rect, Bio & All pitch amp = 25° kinematic cases \\
 \hline
 Gen Test 4 & Rect, Bio & All pitch amp = 40° kinematic cases \\
 \hline
 Gen Test 5 & Rect, Bio, Pt4 & All 2 Hz kinematic cases
 All rect fin cases \\
 \hline
 Gen Test 6 & Rect, Bio, Pt4 & All 2 Hz kinematic cases
 All bio fin cases \\
 \hline
\end{tabular}
\end{table}%

Networks are trained on the subset of data specified for each generalizability test. The overall mean MSE is calculated based on the model’s predictions across all flapping cycles for the excluded kinematic-shape settings. This value is compared with the mean MSE achieved by a model that was trained on the entirety of rectangular and bio fin data. The goal of this comparison is to quantify the deterioration of prediction quality upon the exclusion of data.

\subsection{Neural Network Training}

The networks were trained on data including all provided kinematic-shape settings. For each kinematic-shape setting, one randomly selected flapping cycle from each of the 16 experimental runs was incorporated into the reduced data. An 80/20 training validation split was applied to the reduced data. 

Hyperparameter tuning was run on our baseline models containing a single categorical parameter for fin shape. The baseline models were tuned through a random search method by generating 200 models with hyperparameters within the ranges specified in Table \ref{tab:hyperparameters}. The tuning resulted in the following DNN hyperparameters: layers = 3, nodes per layer = 250, dropout \% = 0.005, batch size = 32, learning rate = 0.007. LSTM hyperparameters are as follows: hidden units = 200, dropout \% = 0.005, batch size = 4, learning rate = $10^{-7}$. The Adam optimizer was used for both networks. DNNs were run for 350 epochs while LSTMs were run for 200 epochs. 

\begin{table}[ht]
\footnotesize
\centering
 \caption{Hyperparameter settings used for DNN and LSTM hyperparameter tuning}
 \label{tab:hyperparameters}
 \begin{tabular}{p{5cm}ll}
 \hline
 Hyperparameter & Min Val & Max Val \\  
 \hline
 \textit{DNN Hyperparameters} & &  \\ 
 layers & 2 & 5 \\
 nodes per layer & 50 & 350 \\
 dropout \% & 0 & 0.2 \\
 batch size & 4 & 64 \\
 learning rate & $10^{-3}$ & $10^{-2}$  \\
 \textit{LSTM Hyperparameters} & & \\ 
 hidden units & 150 & 500 \\
 dropout \% & 0 & 0.2 \\
 batch size & 4 & 64 \\
 learning rate & $10^{-9}$ & $10^{-4}$ \\
\end{tabular}
\end{table}%

\section{Results}

\subsection{Baseline Neural Networks}

The inputs of our baseline DNN and LSTM models consisted of the stroke-related parameters in Table 1 as well as a categorical value for fin shape, which was set to -1 for the pt4 fin, 0 for the rectangular fin, and 1 for the bio fin. The stroke state parameter was excluded for the LSTM: since each LSTM training example represents a full stroke cycle rather than a single point within the cycle, differentiation between upstroke and downstroke can be learned and does not need to be provided. 

The baseline DNN was trained using one stroke cycle from each experimental run and yielded an MSE of 0.068 across the provided data. Almost all of this error is accounted by the experimental data noise given the mean thrust variance of 0.067. The baseline LSTM yielded an MSE of 0.113; with each run, this model must address a more difficult problem of generating the entire thrust profile at once as opposed to predicting thrust at a given point within the flapping cycle. 

\begin{figure}[ht]
  \centering
	\includegraphics[width=0.8\linewidth]{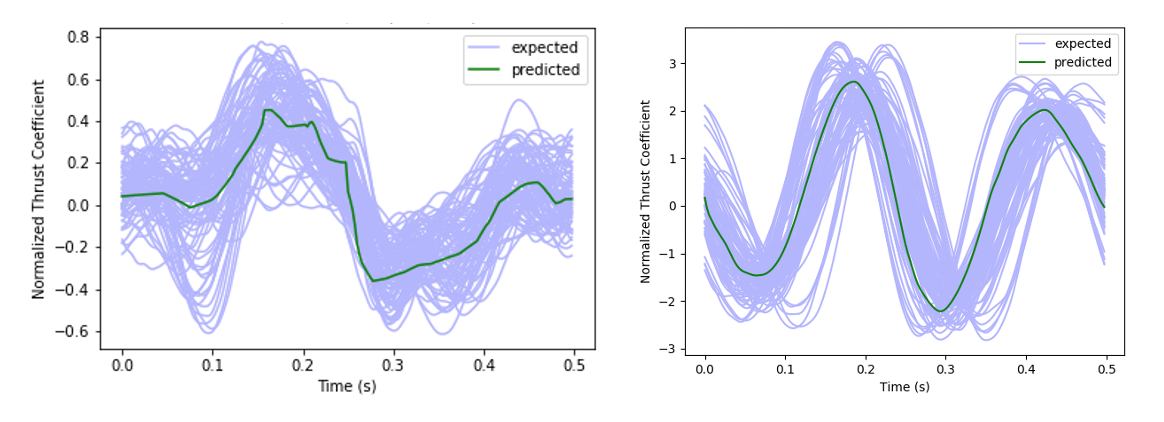}
	\caption{
    Example predicted thrust profiles with the highest MSE for baseline DNN (left) and LSTM (right) trained on complete data. Flapping cycles from experimental data are shown in blue, and the predicted results are shown in green.  
	}
	\label{fig:baseline_cd_prof}
\end{figure}

Both models effectively learn the thrust profiles for all provided kinematic-shape settings. Figure \ref{fig:baseline_cd_prof} demonstrates that both models still generate accurate thrust predictions for their highest MSE predictions. In both cases, the predicted thrust falls within the range of experimental thrust values achieved at every point in time. The LSTM model generates smoother thrust prediction curves since this model maintains a memory of prior predictions and inputs which are used to inform subsequent predictions.

Both networks were then trained based on a subset of the full data set specified by the kinematics-based generalizability tests 1-4. For both networks, an increase in error from the full-data training run is evident across all four generalizability tests (Table \ref{tab:baseline_gen_test}). The baseline LSTM achieved an average MSE of 0.378 across all 4 generalizability tests, slightly outperforming the baseline DNN that had an average MSE of 0.420. 

\begin{table}[ht]
\footnotesize
\centering
 \caption{Baseline DNN and LSTM performance on generalizability tests}
 \label{tab:baseline_gen_test}
 \begin{tabular}{p{2.3cm}p{2.5 cm}p{2.5 cm}}
 \hline
 Test Name & MSE for excluded \newline settings (DNN) & MSE for excluded \newline settings (LSTM)  \\  
 \hline
 Gen Test 1 & 0.289 & 0.222 \\
 Gen Test 2 & 0.671 & 0.534 \\
 Gen Test 3 & 0.452 & 0.466 \\
 Gen Test 4 & 0.268 & 0.288 \\
 \textbf{Gen Test 1-4 Avg} & \textbf{0.420} & \textbf{0.378} \\
\end{tabular}
\end{table}%

 Generalizability test 2, which excludes 15° pitch amplitude data during neural network training, resulted in higher MSEs. The 0° pitch amplitude case does not create an effective stroke and consequentially consists of close to zero thrust noisy flapping cycles; as a result, it becomes harder to generalize to the 15° pitch amplitude case using the two nearest provided pitch amplitudes—0° and 25°--which helps explain the increase in MSE.  
 
\begin{figure}[ht]
  \centering
	\includegraphics[width=0.8\linewidth]{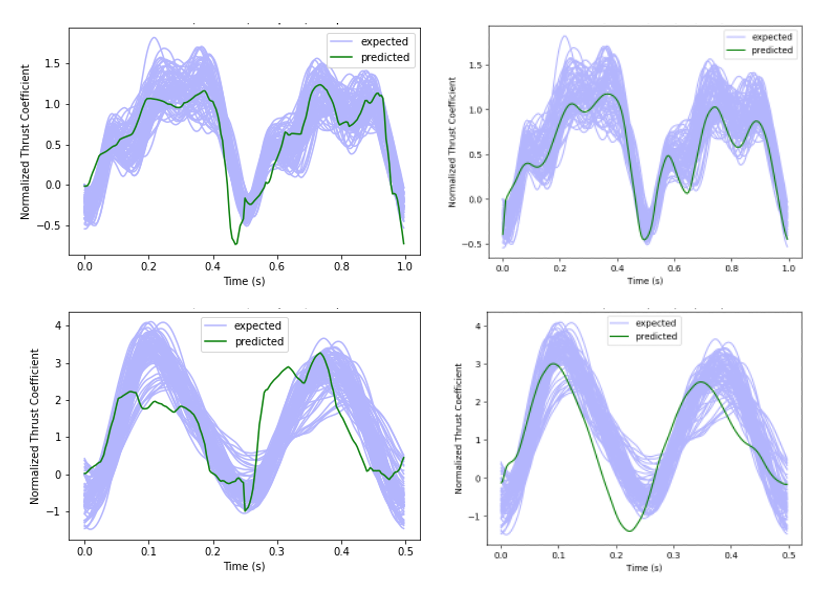}
	\caption{
     Example predicted thrust profiles from generalizability test 3 (excluded of pitch amplitude = 25° data). Among the generated thrust profiles involving interpolated kinematics, the lowest and highest MSE profiles are illustrated on the top and bottom graphs respectively for both the baseline DNN (left) and LSTM (right).
	}
	\label{fig:baseline_gen_prof}
\end{figure}

 Note that both excluded data models generate reasonable interpolated thrust profiles. In the lowest-MSE thrust profile predictions from our baseline models for generalizability test 3, both models successfully detected more subtle peaks within the flapping cycle, such as those located at approximately t = 0.75 and t = 0.85 (Figure \ref{fig:baseline_gen_prof}). As seen from the bottom graphs, the highest-MSE predictions generated by the baseline DNN and LSTM during this test still reasonably capture the two major peaks from the thrust profile both in terms of normalized thrust coefficient magnitude and the location of the peaks within the flapping cycle. Although the worst-case LSTM prediction demonstrates a more accurate peak capture compared to the worst-case DNN prediction, both models demonstrate an ability to interpolate kinematics.

\subsection{Fin Parameterization Neural Networks}

Our fin parameterization neural neural networks (DNN-FP and LSTM-FP) expand upon the baseline models through the addition of 30 fin shape-related inputs. These 30 inputs capture the “skeleton” of the fin shape as described earlier. The fin is partitioned into 10 equal-area segments, from which 10 centers of mass are computed. The 30 inputs correspond to the 3D locations of these centers of mass. 

Training without the exclusion of kinematic-shape settings led to an MSE of 0.085 for DNN-FP and 0.132 for LSTM-FP; these errors are only slightly above our baseline model performance. Both models were then trained based on generalizability tests 1-6, and the resulting MSEs for excluded settings were calculated (Table \ref{tab:FP_gen_test}). Across the tests for kinematic interpolation (generalizability tests 1-4), DNN-FP yielded an average MSE that was 0.063 lower than the baseline DNN while LSTM-FP resulted in an average MSE that was only 0.087 higher than the baseline LSTM. Therefore, both networks were able to maintain their performance in kinematic interpolation despite the addition of new geometry-related inputs.

\begin{table}[ht]
\footnotesize
\centering
 \caption{DNN-FP and LSTM-FP performance on generalizability tests}
 \label{tab:FP_gen_test}
 \begin{tabular}{p{2.3cm}p{2.5 cm}p{2.5 cm}}
 \hline
 Test Name & MSE for excluded \newline settings (DNN) & MSE for excluded \newline settings (LSTM)  \\  
 \hline
 \textit{Kinematics tests} & & \\
 Gen Test 1 & 0.227 & 0.583 \\
 Gen Test 2 & 0.557 & 0.664 \\
 Gen Test 3 & 0.385 & 0.307 \\
 Gen Test 4 & 0.257 & 0.304 \\
 \textbf{Gen Test 1-4 Avg} & \textbf{0.357} & \textbf{0.465} \\
 \textit{Geometry tests} & & \\
 Gen Test 5 & 0.616 & 0.783 \\
 Gen Test 6 & 0.176 & 0.318 \\
 \textbf{Gen Test 5-6 Avg} & \textbf{0.396} & \textbf{0.556} \\
\end{tabular}
\end{table}%

In terms of predicting unseen geometries, DNN-FP and LSTM-FP both reached lower MSEs for bio fin shape prediction in generalizability test 6 compared to the rectangular fin shape prediction in generalizability test 5. These results are surprising since the rectangular fin acts as an approximate interpolation of the bio and pt4 fins by regional COMs (Figure \ref{fig:fin_COMs}) while the bio fin shape must be extrapolated from the rectangular and pt4 fins. 

\begin{figure}[ht]
  \centering
	\includegraphics[width=0.8\linewidth]{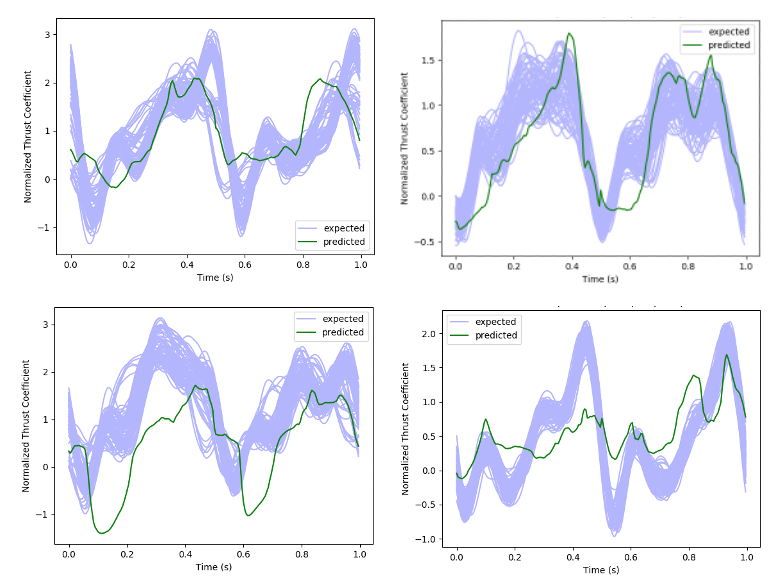}
	\caption{
     Example predicted thrust profiles from DNN-FP for unseen rectangular (left) and bio (right) fin geometries. The lowest and highest MSE cases are illustrated in the top and bottom graphs respectively.
	}
	\label{fig:FP_gen_prof_DNN}
\end{figure}

\begin{figure}[ht]
  \centering
	\includegraphics[width=0.8\linewidth]{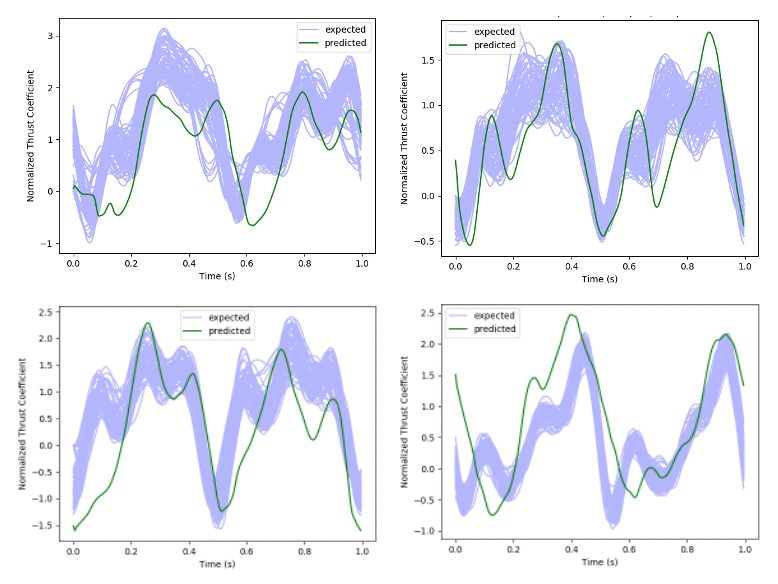}
	\caption{
     Example predicted thrust profiles from LSTM-FP for unseen rectangular (left) and bio (right) fin geometries. The lowest and highest MSE cases are illustrated in the top and bottom graphs respectively.
	}
	\label{fig:FP_gen_prof_LSTM}
\end{figure}

Example DNN-FP and LSTM-FP thrust profile predictions for unseen geometries are shown in Figure \ref{fig:FP_gen_prof_DNN} and Figure \ref{fig:FP_gen_prof_LSTM} respectively. Both networks successfully predicted the magnitude and location of the two dominant peaks in the rectangular and bio fin lowest-MSE cases. However, even during these lowest-MSE cases, the models missed more subtle trends such as the peak at 0.2s for the illustrated rectangular fin thrust profile. For the highest-MSE case in each generalizability test, LSTM-FP accurately depicted the dominant peaks present in the flapping cycle while the DNN-FP was unable to fully realize the first major peak. Overall, from figures \ref{fig:FP_gen_prof_DNN} and \ref{fig:FP_gen_prof_LSTM}, we see that DNN-FP and LSTM-FP demonstrate some predictive power to unseen fin shapes in both fin geometry exclusion cases. 

\subsection{Reduced Kinematic Fin Parameterization Neural Networks}

The proposed fin parameterization captures the dynamic motion of the fin: the 3D locations of the centers of mass change based on the current stroke and pitch angles. As a result, our fin parameterization could serve as a replacement for 4 of our 6 kinematic parameters, namely stroke angle, pitch angle, stroke amplitude, and pitch amplitude. We created two new networks (DNN-RFP and LSTM-RFP) with a simplified input space that included only the flap frequency and stroke state along with the 30 fin parameterization values. The removal of 4 kinematic parameters could force the network to more effectively learn the relationship between the fin geometry parameters and thrust. 

\begin{figure}[ht]
  \centering
	\includegraphics[width=0.8\linewidth]{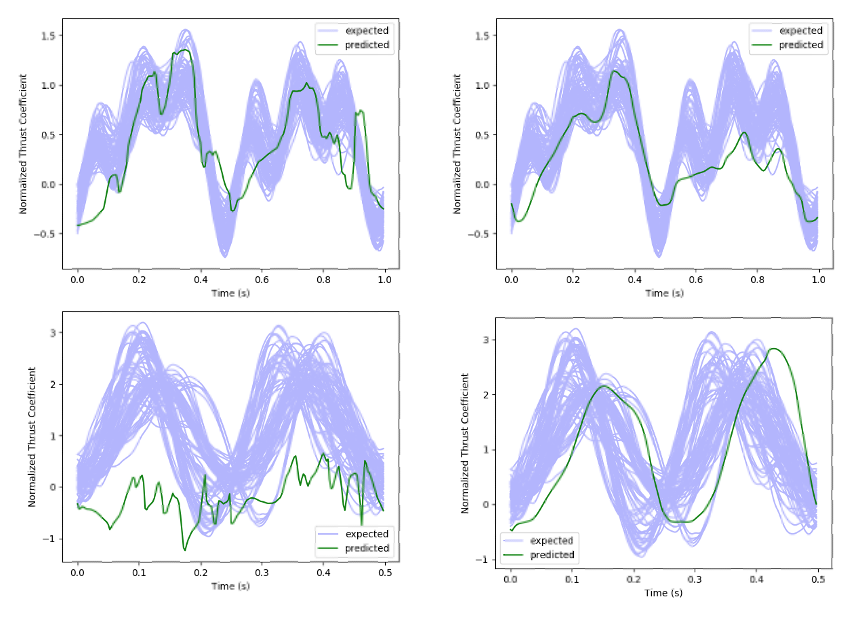}
	\caption{
     Deteriorated predicted thrust profiles from generalizability test 2 (excluded of pitch amplitude = 15° data). Among the generated thrust profiles involving interpolated kinematics, the lowest and highest MSE profiles are illustrated on the top and bottom graphs respectively for both DNN-RFP (left) and LSTM-RFP (right).
	}
	\label{fig:RFP_gen_prof}
\end{figure}

When all kinematic-shape settings were included for training, DNN-RFP and LSTM-RFP reached a MSE of 0.15 and 0.139 respectively. Since both models learned the thrust profiles across provided kinematic-shape settings, the proposed fin parameterization acts as a viable substitute for the information provided by the stroke and pitch angle and amplitude inputs. However, exclusion of these 4 kinematic inputs resulted in a reduced ability for the networks to generalize to new kinematics. The average MSE for excluded settings among the kinematic interpolation generalizability tests increased from 0.357 to 0.855 and 0.465 to 0.703 for the DNN and LSTM-based model respectively (Table \ref{tab:RFP_gen_test}). This deterioration can be seen graphically from the generalizability test 2 predictions shown in Figure \ref{fig:RFP_gen_prof}. In the lowest-MSE case, LSTM-RFP underestimates the magnitude of the second major peak. In the highest-MSE case, LSTM-RFP inaccurately predicts the locations of both dominant peaks while DNN-RFP is unable to predict either peak.  

\begin{table}[ht]
\footnotesize
\centering
 \caption{DNN-RFP and LSTM-RFP performance on generalizability tests}
 \label{tab:RFP_gen_test}
 \begin{tabular}{p{2.3cm}p{2.5 cm}p{2.5 cm}}
 \hline
 Test Name & MSE for excluded \newline settings (DNN) & MSE for excluded \newline settings (LSTM)  \\  
 \hline
 \textit{Kinematics tests} & & \\
 Gen Test 1 & 0.711 & 0.476 \\
 Gen Test 2 & 1.203 & 0.662 \\
 Gen Test 3 & 0.838 & 0.268 \\
 Gen Test 4 & 0.716 & 0.407 \\
 \textbf{Gen Test 1-4 Avg} & \textbf{0.867} & \textbf{0.453} \\
 \textit{Geometry tests} & & \\
 Gen Test 5 & 0.624 & 0.824 \\
 Gen Test 6 & 0.777 & 0.384 \\
 \textbf{Gen Test 5-6 Avg} & \textbf{0.701} & \textbf{0.604} \\
\end{tabular}
\end{table}%

In addition, the removal of kinematic-based parameters did not result in a decrease in MSE across the geometry generalizability tests in either case: DNN-RFP and LSTM-RFP yielded an average geometry test MSE of 0.701 and 0.604 while DNN-FP and LSTM-FP reached an MSE of 0.396 and 0.556 (Table \ref{tab:RFP_gen_test}). Therefore, the reduced networks were unable to more effectively learn the relationship between the 30 fin shape parameters and thrust. Additionally, DNN-RFP performed worse than DNN-FP across the kinematic generalizability tests (the networks yielded average MSEs of 0.867 and 0.357 respectively), so the inclusion of explicit parameters detailing stroke and pitch improves kinematic interpolation for our DNN model. 

\subsection{Reduced-Order Fin Parameterization Neural Networks}

To simplify our fin parameterization, we reduced the fin geometry input space from 30 parameters to 4 parameters that capture the vast majority of the input space. Principal component analysis with a weighting procedure is used for this dimensionality reduction as described in our methods. The resulting DNN-WFP and LSTM-WFP networks incorporate 4 fin shape parameters in addition to the kinematic parameters described in Table \ref{tab:parameters}. 

With the inclusion of all kinematic-shape settings during training, DNN-WFP yielded an MSE of 0.071, and LSTM-WFP yielded an MSE of 0.152. These low MSEs demonstrate how our simplified fin parameterization is sufficiently descriptive to reflect the differences between the different fin geometries. The average MSEs during kinematic interpolation were 0.855 and 0.703 for DNN-WFP and LSTM-WFP respectively; these networks were outperformed by their 30 fin shape parameter counterparts, which achieved average MSEs of 0.357 and 0.465 (Table \ref{tab:WFP_gen_test}). 

\begin{table}[ht]
\footnotesize
\centering
 \caption{DNN-WFP and LSTM-WFP performance on generalizability tests}
 \label{tab:WFP_gen_test}
 \begin{tabular}{p{2.3cm}p{2.5 cm}p{2.5 cm}}
 \hline
 Test Name & MSE for excluded \newline settings (DNN) & MSE for excluded \newline settings (LSTM)  \\  
 \hline
 \textit{Kinematics tests} & & \\
 Gen Test 1 & 1.087 & 1.065 \\
 Gen Test 2 & 0.925 & 0.733 \\
 Gen Test 3 & 0.481 & 0.495 \\
 Gen Test 4 & 0.925 & 0.52 \\
 \textbf{Gen Test 1-4 Avg} & \textbf{0.855} & \textbf{0.703} \\
 \textit{Geometry tests} & & \\
 Gen Test 5 & 0.94 & 0.674 \\
 Gen Test 6 & 0.393 & 0.366 \\
 \textbf{Gen Test 5-6 Avg} & \textbf{0.667} & \textbf{0.52} \\
\end{tabular}
\end{table}%

\begin{figure}[ht]
  \centering
	\includegraphics[width=0.8\linewidth]{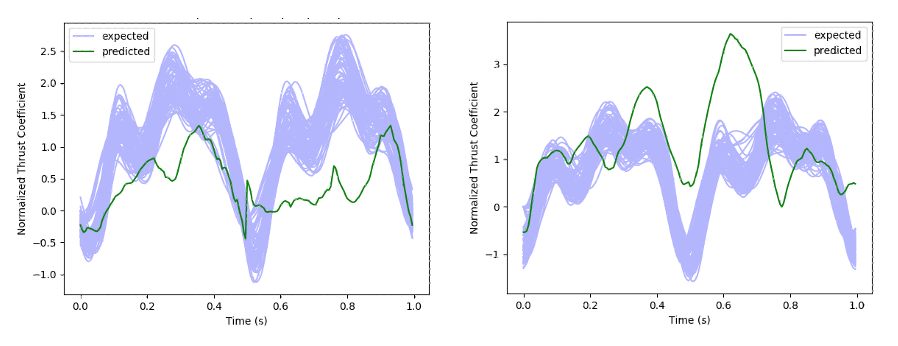}
	\caption{
     Example predicted thrust profiles from DNN-WFP for unseen rectangular (left) fin geometry. The lowest and highest MSE cases are illustrated in the left and right graphs respectively.
	}
	\label{fig:WFP_gen_prof_DNN}
\end{figure}

Across the geometry generalizability tests, LSTM-WFP maintained an average MSE of 0.52, which is similar to the average MSE of 0.556 from LSTM-FP. However, DNN-WFP performed worse in both geometry tests compared with DNN-FP, reaching an average MSE of 0.667 while DNN-FP reached an average MSE of 0.396. This deterioration is evident from the inability of DNN-WFP to correctly predict the magnitude of the dominant peaks in the lowest-MSE (best-performing) rectangular fin case (Figure \ref{fig:WFP_gen_prof_DNN}).  Therefore, the dimensionality reduction for fin parameterization did not facilitate an easier learned relationship between the fin parameterization values and thrust.

\section{Discussion}

We experimented with neural network-based models that used a flapping fin system's fin geometries and kinematics for fast UUV thrust prediction. Two main architectures were proposed: our DNN-based models predict instantaneous thrust while our LSTM-based models allow for single-cycle thrust profile predictions. The LSTM-based approach maintains a memory of past inputs and output thrusts at the expense of creating a more complex time-series problem rather than a point prediction problem.  

Our baseline models with a categorical fin shape parameter demonstrate reasonable interpolation to unseen kinematics. With the incorporation of COM-based fin \add[JL]{shape} parameterization, our DNN-FP and LSTM-FP networks were able to generalize to unseen fin shapes while maintaining the baseline performance for kinematic interpolation. \add[JL]{Unlike prior proposed fin shape parameterization methods, parameters in our approach account for fin position in addition to fin shape, leading to dynamic values that change over the course of a flapping cycle. This approach aims to overcome the point-based domain coverage achieved by a static fin parameterization method, which could become problematic given our input data consists of a small set of fin shapes. Future work will involve a direct comparison between dynamic and static fin shape parameterization methods.}

\change[JL]{The}{Our} fin shape parameter\add[JL]{s} \remove[JL]{ inputs} implicitly include kinematic information such as the current stroke and pitch angle. However, the removal of these explicit kinematic parameters results in decreased kinematic and fin geometry generalizability for the DNN model. Additionally, this removal does not improve the ability of the LSTM model to generalize to unseen fin shapes, so reducing the number of kinematic parameters does not reinforce a better learned relationship between the fin shape parameters and thrust.

While most of the information within our 30-value fin parameterization can be captured using far fewer parameters, such a dimensionality reduction does not help the model relate the fin parameterization to output thrust: the modified DNN demonstrated lower generalizability to unseen fin shapes while the modified LSTM did not show an improved performance from LSTM-FP. Additionally, this modification resulted in worse kinematic interpolation for both models.

Overall, one network architecture did not clearly outperform the other with respect to kinematic and fin shape generalizability. However, LSTM models produced smoother thrust profiles since our DNNs do not maintain a memory of prior predictions to inform outputs. LSTM models were also more resilient to changes in input: the LSTMs generally maintained performance in terms of kinematic and fin shape generalizability with the removal of kinematic parameters and the introduction of fin shape parameter dimensionality reduction.  

The incorporation of new data for a larger array of distinct fin geometries could lead to more effective fin shape generalizability in the future. Meanwhile, we aim to modify our existing models to predict thrust for the back fin of a 4 pectoral fin system. We will also work towards devising new fin parameterization strategies, including strategies that account for the deformity of flexible fins in the fin parameterization.\add[KV]{ Our plan includes exploring non-linear models to reduce the input space parameters, in lieu of the linear POD method, and arrive at a reduced order representation that is sufficiently general and practical for use in control system algorithms} We have achieved promising results in terms of kinematic interpolation and generalizability to new fin shapes; future iterations of our thrust prediction models could be used to generate potential high-thrust fin geometries, and these models could be embedded into UUV control systems.


\fontsize{9.0pt}{10.0pt}
\bibliographystyle{aaai}
\selectfont
\bibliography{ms}

\end{document}